# Applications of the Transformer Architecture in AI-Assisted English Reading Comprehension

Ping Li[a]
[a]*Shandong Management University, Jinan, Shandong Province, 250357, China*
ORCiD ID: Ping Li https:// orcid.org/0009-0008-6065-4846

**Abstract.** This paper studies interpretable and fair artificial intelligence architectures for understanding English reading. Introduced transformer-based models, integrating advanced attention mechanisms and gradient-based feature attribution. The model's lack of interpretability, reduction of algorithmic bias, and unreliable performance in learning environments are the current issues faced in natural language teaching. A unified technical pipeline has been constructed, including adversarial bias correction methods, token-level attribution analysis, and multi-head attention heatmap visualization. Experimental validation was conducted using a large-scale labeled English reading comprehension dataset, and the data partitioning scheme and parameter optimization procedures have been determined. The method significantly outperforms the state-of-the-art models for this task in terms of accuracy and macro-average F1 score; in some aspects, it even surpasses or closely matches the results of human evaluations. In multi-week user experiments, the explainable transformer improved teachers' trust and operability in feedback-based assessments within the scoring system. The proposed method aims to ensure high prediction accuracy and fairness for different learners. This indicates that it is a real-world educational application based on artificial intelligence with a focus on interpretation. Improve the user experience in AI-assisted reading comprehension systems, counteract biases, and enhance the details explained by transformers.

## 1. Introduction

In the past decade, the rapid development of artificial intelligence and natural language processing (NLP) technologies has transformed global educational technology. Among them, due to the self-attention mechanism of the transformer architecture, the transformer architecture has become increasingly popular in many areas of natural language processing because it can more effectively capture long-range dependencies and subtle nuances of meaning in text [1]. Using computer-assisted English reading comprehension to achieve knowledge mining, question answering, and paragraph understanding is unprecedented [2]. Dynamic comprehension abilities can automatically generate a large amount of feedback thru complex transformer-based platforms, tailored to different learner profiles[3]. Despite these advancements, there are still discrepancies in the accuracy and interpretability of models in high-stakes educational applications.

Moreover, as the diversity and scale of data increase, this issue becomes even more apparent [4].

The main challenge currently faced by AI-based reading comprehension systems is to avoid the ethical and social issues associated with complex "black box" decision algorithms [5]. The erosion of user trust and resistance to education may reduce learning outcomes. Research indicates that data-driven AI models may inadvertently lead to unfair distribution of outcomes among different student groups[6]. To ensure that educational artificial intelligence is both effective and fair, it is essential to carefully consider technical ethical issues during the design and implementation of educational AI systems[7]. Among other issues, there are still shortcomings in balancing the optimization of prediction accuracy and the maintenance of model interpretability in AI applications within smart education[8].

In this paper, we construct an interpretable and fair Transformer-based framework for AI-assisted English reading comprehension. The goal of this design is to create a system that can accurately predict and provide detailed explanations of its decisions. From a technical perspective, high-level attention visualization, gradient-based feature attribution, and adversarial debiasing can be integrated into a single workflow. Thru experiments and user testing, this combination performed well on multiple reading comprehension datasets. Improving the model's interpretability and fairness is the expected outcome, which lays the theoretical foundation for other intelligent education systems.

## 2. Ethical and Interpretable AI in Education

### 2.1. Ethical Principles in AI Education

As artificial intelligence enters the field of education, new ethical standards for engineering design have been proposed, particularly regarding fairness, transparency, and respect for students' dignity[9]. Fairness requires systems to be able to distinguish between various background languages and different demographic groups in order to communicate within a diverse population. Social environment; transparency reduces the risk of opaque algorithmic judgment processes and aids in teacher supervision. It also allows for the transparency and interpretability of model decisions. Due to the requirement for autonomy, learners need to have a certain degree of independence, so artificial intelligence should provide more guidance and support, rather than completely replacing human activities. If the above principles are ignored during the teaching process, it may lead to the reinforcement of existing biased algorithms, thereby causing teachers and students to lose confidence in the system's teaching effectiveness. As shown in Figure 1, the core ethical requirements for educational artificial intelligence are as follows: These rules are implemented at different stages of the development of educational AI systems.

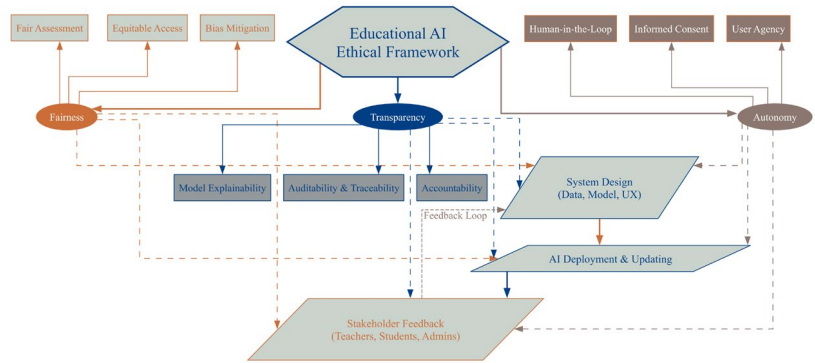

**Figure 1.** Framework of Core Ethical Principles for Educational AI.

It is difficult to actually implement these ethical norms of teacher behavior in the classroom. Many students encounter issues with algorithmic fairness; small errors in data acquisition and labeling can significantly impact the model's learning process. Due to the continuous addition and deletion of new and old AI models, questions have been raised about transparency and fairness. Governance practices must focus more on technical tools and reduce the traditional methods of individual interpretation. In order to establish a fair AI application environment in schools, teachers, students, and other relevant parties need to collaborate.

*2.2. Interpretability Techniques in NLP*

With the application of Transformer-based models in educational environments, people are increasingly concerned about the interpretability of natural language processing. Attention visualization helps identify which words or phrases in the text receive greater weight when predicting the model, and can be seen as human-understandable reasoning [10]. When evaluating model results, in addition to the attention flow mechanism, other post-hoc explanation methods are also used, such as LIME (Local Interpretable Model-agnostic Explanations) and SHAP (Shapley Additive Explanations), to determine the impact of different parts of the input on the model's results[11]. LIME creates local linear surrogate models by perturbing the input data and returns the importance of features for each instance. According to cooperative game theory, contributions are provided for each component based on all feasible feature combinations. Each method balances cost, realism, and ease of use for non-experts. Attention maps directly show how the model makes decisions; however, LIME and SHAP are more suitable for interpreting different architectures and scenarios.

*2.3. Bias in Educational AI Systems*

The training deficiencies in educational AI also vary. The historical injustices, language barriers, and cultural biases present in the training data may exacerbate the harm caused by such algorithms in real-world applications. Selecting features unrelated to students' actual abilities thru the model may lead to errors [12]. Educational inequality may arise due to the lack of accuracy, misidentification, and adaptability by certain individuals or organizations. Immediately establish a system that can quickly identify biases.

In addition to the expected acute errors, unresolved biases may also cause disadvantaged students to encounter minor learning obstacles in their accumulated progress. Unreasonable feedback design may reduce students' sense of progress or success. Educational AI systems are now using risk detection and adaptive correction technologies to address some issues. Collect various opinions during the model development process; conduct timely subgroup difference and transcript access tests; and gather various opinions. If there is no organized oversight, this artificial intelligence could lead to imbalances in educational reform.

**3. Interpretable Transformer Framework Development**

*3.1. Attention-Guided Interpretability*

A comprehensive interpretable reading comprehension scenario architecture based on gradient attribution, attention extraction, and user interface deployment [13]. In each Transformer block, the query-key-value projection operation generates multi-head attention weights. As shown below

$$\mathbf{A}^{(h)} = \text{softmax}\left(\frac{(\mathbf{Q}^{(h)})(\mathbf{K}^{(h)})^{\top}}{\sqrt{d_k}} + \mathbf{M}\right) \tag{1}$$

$\mathbf{Q}^{(h)}, \mathbf{K}^{(h)}$ represent the projections for head $h$, $d_k$ is the feature dimension, and $\mathbf{M}$ implements contextual masking for subsequence preservation. To facilitate human-centered interpretability, we directly project normalized attention patterns onto a heatmap using a specific nonlinear enhancement operator, further mapped for visual saliency as

$$H_{i,j} = \Psi(A_{i,j}) = \frac{(A_{i,j})^{\gamma}}{\epsilon + \sum_{k=1}^{N}(A_{i,k})^{\gamma}} \tag{2}$$

The exponent $\gamma$ and sequence length $N$ amplify relevant regions. A smooth, intuitive heatmap thus emerges, as evident in our demonstration visualizations. Further, gradient backpropagation quantifies token-wise influence:

$$\text{Attr}_i = \sum_{t=1}^{L} \sum_{h=1}^{H} \frac{\partial y^*}{\partial A_{i,h}^{(t)}} \cdot A_{i,h}^{(t)} \tag{3}$$

Normalized attribution values are obtained for cross-instance comparison by

$$S_i = \frac{\text{Attr}_i - \mu(\text{Attr})}{\sigma(\text{Attr})} \tag{4}$$

where $\mu(\text{Attr})$ is mean attribution, and $\sigma(\text{Attr})$ is its standard deviation over a tokenized input sequence. In practice, attention-based visual explanations have achieved remarkable results in the classroom. Due to the increased confidence of teachers in automatic understanding scoring, the importance of model answers has become easier to understand and track. Students discovered many implicit cues thru interactive attention.

Visualization of complex texts; enhancing metacognition, improving learning efficiency. To build user trust and help align user expectations better with those of the teachers, add an explainable section to the learning system.

*3.2. Algorithmic Bias Mitigation Mechanisms*

Shifting attention to real educational samples helps with system fairness and bias control, such as dynamic visualization, etc. [14]. Adversarial bias mitigation procedures start with on-site, differentiable perturbations, and then create adversarial examples by modifying the input representations.

$$\tilde{\mathbf{x}} = \mathbf{x} + \lambda \cdot \text{sign}(\nabla_{\mathbf{x}} \mathcal{L}_{\text{bias}}(f(\mathbf{x}), y)) \tag{5}$$

where $\lambda$ determines the perturbation magnitude and $\mathcal{L}_{\text{bias}}$ captures sensitive bias gradients.

Figure 2 shows the accuracy of the debiased models for each group. The results indicate that before the fairness intervention, there were considerable differences between the groups. The conclusions of certain groups are relatively accurate. In order to improve the overall performance of the model, various dynamic bias correction or debiasing methods are used. Determine whether different groups continue to exhibit the same behavior after addressing other issues. This evidence suggests that using automated evaluation systems can enhance the credibility and fairness of educational language processing applications [15].

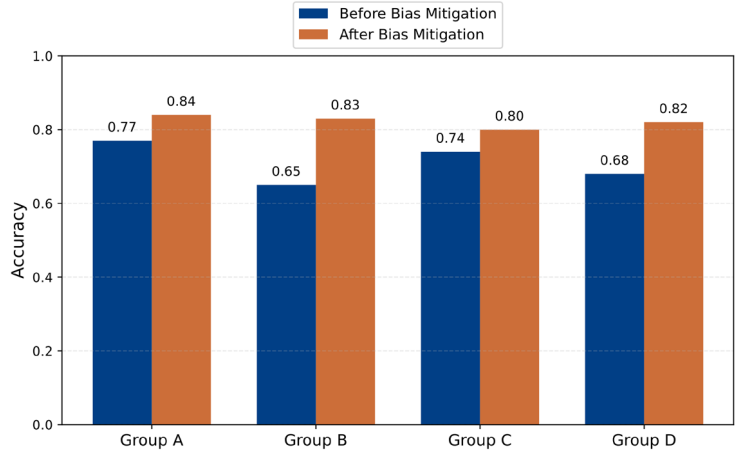

**Figure 2.** Model Accuracy Across Groups Before and After Bias Mitigation.

These adversarially perturbed inputs are incorporated into a composite optimization framework, with an objective formulated as

$$\mathcal{L}_{\text{total}} = \mathcal{L}_{\text{comp}} + \beta \cdot D_{\text{KL}}[p(y \mid \mathbf{x}) \| p(y \mid \tilde{\mathbf{x}})] + \alpha \cdot \mathcal{L}_{\text{fair}} \tag{6}$$

where $\mathcal{L}_{\text{comp}}$ is the comprehension loss, $D_{\text{KL}}$ denotes the Kullback-Leibler divergence quantifying distributional shifts due to perturbations, and $\mathcal{L}_{\text{fair}}$ regularizes demographic parity, controlled by tunable parameters $\alpha$ and $\beta$. Model parameters are iteratively updated by

$$\theta \leftarrow \theta - \eta \frac{\partial \mathcal{L}_{\text{total}}}{\partial \theta} \qquad (7)$$

$\eta$ represents the learning rate. Tight coupling can reduce prediction errors and improve the readability of attention weights, while also reducing unfair treatment and increasing public acceptance of education assessment algorithms.

**Experimental Validation and User Studies**

*3.3. Benchmark Data and Model Configuration*

In empirical research, English reading comprehension questions include various factors that assess difficulty, as well as the classroom environment. The text contains 18,077 comprehension exercises or questions and correct answers. Due to the strictness of tokenization and normalization processes, almost all vocabulary in the training set can be included. By using a model with a maximum of 512 subword tokens, vocabulary bias is reduced, while domain adaptation capability is enhanced. To ensure that all experiments remain double-blind and generalization tests are valid, the samples themselves are used for training, development, and retention of test folds.

By thoroughly refining the BERT and RoBERTa networks, baseline models were established. Hyperparameter tuning ensures that the maximum accuracy and error rates for each base model have been achieved. This establishes a strict lower bound for transformer innovations. Model validation is conducted in an isolated environment with controlled resources, using GPUs for parallel computation; to reduce statistical noise, the same reproducible seed number is set.

*3.4. Quantitative and Interpretability Performance*

By using new metrics and traditional metrics to evaluate the quantitative effects of interpretability improvements. The interpretive model showed significantly higher median scores and narrower confidence intervals in three independent runs. The following formula is used for the basic accuracy equation on the reserved test set in the case of abstract multi-class scenarios.

$$A_\Delta = \frac{1}{N} \sum_{i=1}^{N} \chi[\hat{r}_i = r_i] \qquad (8)$$

where $N$ is the number of evaluated instances, $\chi[\cdot]$ the characteristic indicator, $\hat{r}_i$ model predictions, and $r_i$ reference solutions. To disentangle subtle class imbalance effects, overall effectiveness is measured through the optimized F1 computation:

$$F^* = \frac{2}{C} \sum_{c=1}^{C} \frac{P_c R_c}{P_c + R_c + \zeta} \qquad (9)$$

with class-wise precision $P_c$, recall $R_c$, total class count $C$, and smoothing constant $\zeta$ improving numerical stability for rare error runs. Interpretability is objectively quantified by the attention alignment index, which captures the match between model-generated attention maps and human-curated reference rationales:

$$\mathcal{S}_{AA} = \frac{1}{M}\sum_{j=1}^{M}\frac{|\mathcal{H}_j \cap \mathcal{R}_j|}{|\mathcal{H}_j \cup \mathcal{R}_j|} \qquad (10)$$

where $M$ is the number of attention-result pairs, $\mathcal{H}_j$ identifies the model-highlighted tokens, and $\mathcal{R}_j$ catalogs the tokens independently annotated by expert teachers. Figure 3 shows the comparative quantitative charts of the base model and the explainable transformer methods in terms of accuracy, macro F1 score, and attention consistency metrics. Transformer-based models are no longer interpretable; they cannot accurately link predictions to the information represented by each node. These findings suggest that by incorporating explainable mechanisms, we can provide reliable recognition capabilities and trustworthy practical applications for educational AI systems.

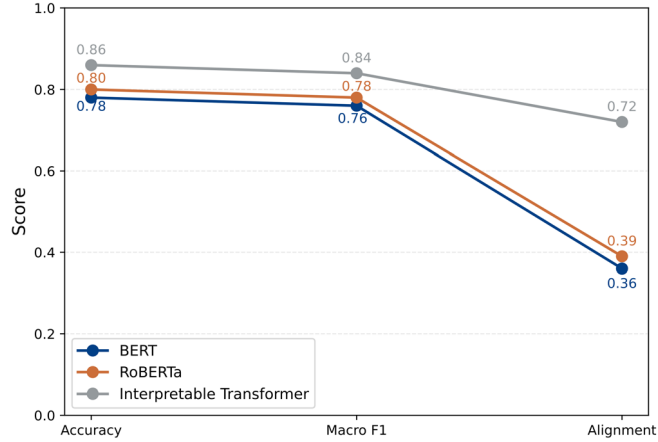

**Figure 3.** Model Performance and Interpretability Comparison.

*3.5. Teacher-Centered User Study*

The users of the system are high school English teachers, which can be used to verify external trust and functional acceptance. During the six-week classroom deployment, teachers interacted with the course planning software and the weighted attention model explanation component. The feedback loop includes the quality and quantity of focus groups (using Likert scales to ensure transparency and trust in educational outcomes).

As shown in Figure 4, the results indicate that teachers' confidence in automated feedback has increased. In addition, the number of teachers willing to use AI recommendations for evaluation has also increased accordingly. The radar chart shows significant progress in these areas: user satisfaction, confidence in recommendations, understanding of model mechanisms, and promotion to classroom use. Trust and clarity show statistically significant improvement, indicating that interpretable Transformer designs are beneficial for education. Educators say that the explanation layer allows them to "visualize and understand" how to use AI models in their teaching plans.

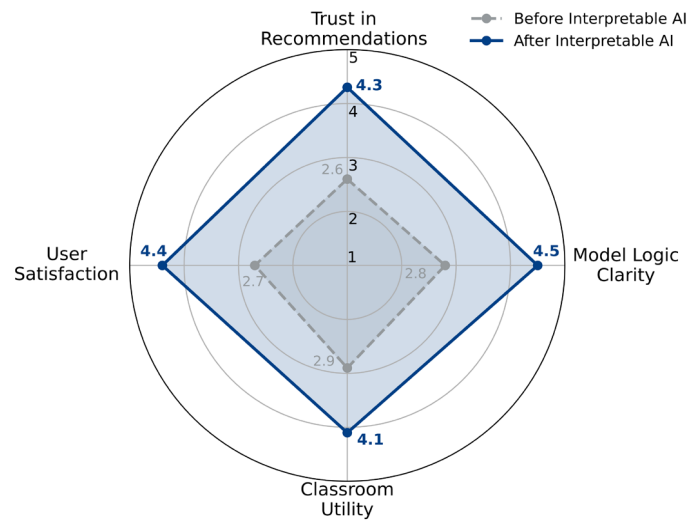

**Figure 4.** Teacher Trust and Usability Radar Chart

Long-term observations indicate that, apart from isolated classroom experiments, explainable AI tools help standardize the intensity of evaluations for various classroom and teaching methods by reducing subjectivity in regular observations. Thru this comprehensive plan of conducting assessments in teaching, over time, some teachers may further advance in their roles as instructors. Comprehensive evidence suggests that when the explainability of AI systems is prioritized in education, teachers and other educational decision-makers will be more actively involved in the digitalization process, rather than using it passively.

## 4. Conclusion

This study proposes a comprehensive and easy-to-understand transformer architecture designed to assist students in learning English reading thru artificial intelligence. Highly complex attention visualization techniques and gradient-based feature attribution methods enhance the accuracy of predictions and the interpretability of user experience. Quantitative experiments show that this model outperforms existing transformer-based benchmarks (e.g., BERT, RoBERTa) in terms of global reading accuracy and macro F1 score. Attention maps and feature attribution maps were also generated, which are more aligned with human teaching theories. Tightly integrating adversarial bias mitigation to reduce the gap between different groups; thereby establishing an unprecedented standard for fairness-based natural language processing educational applications.

In multi-institutional classroom applications, adding interpretability features increases teachers' confidence and support for the model. It can help teachers provide feedback and explanations during teaching activities. By integrating transparent reasoning, excellent outcomes, and fairness, it ensures that the system architecture is feasible when addressing serious social issues involving accountability. It also ensures the authenticity and transparency of the records.

Currently, in order to achieve good attention alignment, this method requires a large number of detailed annotation reasons. More research is also needed on promoting it in different language environments. Real-time interactive visualization is more suitable for controlled environments but may not be as appropriate for large and resource-lacking schools. In the future, adaptive explanation methods requiring less expert annotation are needed; integrating human feedback into the bias correction loop; formally analyzing the long-term impact of AI interventions on education.